\documentclass[letterpaper]{article} 
\usepackage{aaai24}  
\usepackage{times}  
\usepackage{helvet}  
\usepackage{courier}  
\usepackage[hyphens]{url}  
\usepackage{graphicx} 
\usepackage{natbib}  
\usepackage{caption} 
\usepackage{algorithm}
\usepackage{algorithmic}
\usepackage{multirow}
\usepackage{amsmath}

\usepackage{newfloat}
\usepackage{listings}
\DeclareCaptionStyle{ruled}{labelfont=normalfont,labelsep=colon,strut=off} 
\floatstyle{ruled}
\newfloat{listing}{tb}{lst}{}
\floatname{listing}{Listing}
\title{SD-MVS: Segmentation-Driven Deformation Multi-View Stereo 

with Spherical Refinement and EM optimization}
\author{
    Zhenlong Yuan\textsuperscript{\rm 1}, 
    Jiakai Cao\textsuperscript{\rm 1}, 
    Zhaoxin Li\textsuperscript{\rm 2, \rm 3}\thanks{Corresponding Author.}, 
    Hao Jiang\textsuperscript{\rm 1}, 
    Zhaoqi Wang\textsuperscript{\rm 1}
}
\affiliations{
    \textsuperscript{\rm 1}Institute of Computing Technology, Chinese Academy of Sciences
    
    \textsuperscript{\rm 2}Agricultural Information Institute, Chinese Academy of Agricultural Sciences
    
    \textsuperscript{\rm 3}Key Laboratory of Agricultural Big Data, Ministry of Agriculture and Rural Affairs

    yuanzhenlong21b@ict.ac.cn, caojiakai21@mails.ucas.ac.cn, 
    
    cszli@hotmail.com, \{jianghao, zqwang\}@ict.ac.cn
}

\usepackage{bibentry}

\begin{document}

\maketitle

\begin{abstract}
In this paper, we introduce Segmentation-Driven Deformation Multi-View Stereo (SD-MVS), a method that can effectively tackle challenges in 3D reconstruction of textureless areas. We are the first to adopt the Segment Anything Model (SAM) to distinguish semantic instances in scenes and further leverage these constraints for pixelwise patch deformation on both matching cost and propagation. Concurrently, we propose a unique refinement strategy that combines spherical coordinates and gradient descent on normals and pixelwise search interval on depths, significantly improving the completeness of reconstructed 3D model. Furthermore, we adopt the Expectation-Maximization (EM) algorithm to alternately optimize the aggregate matching cost and hyperparameters, effectively mitigating the problem of parameters being excessively dependent on empirical tuning. Evaluations on the ETH3D high-resolution multi-view stereo benchmark and the Tanks and Temples dataset demonstrate that our method can achieve state-of-the-art results with less time consumption.
\end{abstract}

\section{Introduction}

Multi-view stereo (MVS) is a technique that employs images to reconstruct 3D objects or scenes. Its application spans various fields, including autonomous driving \cite{orsingherRevisitingPatchMatchMultiView2022}, augmented reality \cite{caoAccurate3DReconstruction2021}, and robotics \cite{liDenseSurfaceReconstruction2019}.

\par
Recently, PatchMatch-based methods \cite{leibe_pixelwise_2016, xu_multi-scale_2019, PM-RL} exhibits remarkable capabilities in sub-pixel reconstruction for large-scale imagery while being reliable for unstructured image set.
These methods typically initiate by computing the matching cost of fixed patches between images, then proceeding with propagation and refinement for accurate depth estimation. Nonetheless, they typically encounter difficulties in textureless areas where the absence of texture results in unreliable depth estimations. 
To address this issue, several techniques have been introduced, including plane prior \cite{xu_planar_2020}, superpixel-wise planarization \cite{romanoni_tapa-mvs_2019}, epipolar geometry \cite{xu_marmvs_2020}  and confidence-based interpolation \cite{li_confidence-based_2020}. Yet when facing large textureless areas, these methods perform unsatisfactory and leave room for further improvement.

\par
Differently, learning-based methods leverages network to build learnable 3D cost volumes and thereby ameliorating the reconstruction quality. 
Several methods \cite{yao_recurrent_2019, HC-RMVSNet} attempt to employ the gated recurrent unit (GRU) to provide a more rational interpretation in reconstruction, while this often leads to unaffordable time and memory cost.
Others \cite{EPNet} try to utilize residual learning module to refine depth estimates by rectifying the upsampling errors.
Yet, such networks typically lack generalization when facing scenes different from training datasets, posing challenges for their practical application.

\begin{figure}
    \centering
    \includegraphics[width=\linewidth]{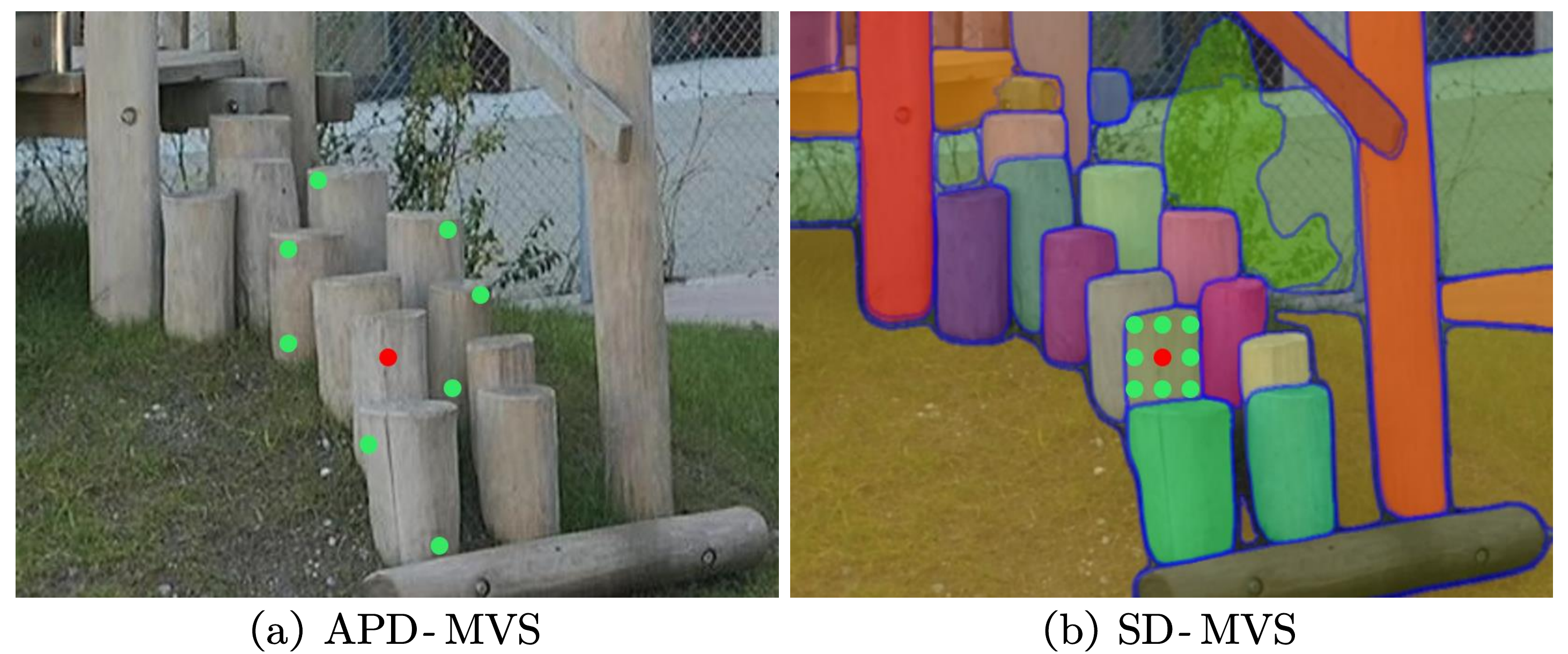}
    \caption{Comparative analysis of patch deformation strategies between APD-MVS and our approach. APD-MVS (a) selects green anchor pixels from pixels characterized by similar colors but may have inconsistent depths to help reconstruct central red pixel, leading to potential inaccuracy. Conversely, our method (b) utilizes neighboring pixels inside the segmentation boundary for reconstruction.
    }
    \label{fig:1}
\end{figure}

\par
Edges in the color image are usually consistent with depth boundaries. Thus, edge information plays a pivotal role in both computation of PatchMatch and construction of 3D cost volumes. 
Nonetheless, problems like shadows and occlusions in complicated scenes tend to weaken the linkage between edge and depth boundaries.
Consequently, several methods \cite{yuesong_wang_adaptive_2023} struggle to harness edge information effectively, often skipping edges and consequently calculating regions with inconsistent depth, leading to detail distortion, as shown in Fig. \ref{fig:1}.
Additionally, certain superpixel segmentation approaches \cite{fink_plane_2019} face challenges in precisely segmenting edges and lack semantic information to broaden receptive field.
Differently, as an instance segmentation model, the Segment Anything Model (SAM) \cite{kirillov2023segany} can subtly mitigates the aforementioned disturbances, thereby segmenting instances with different depths across diverse scenes.

\par
Therefore, we introduce SD-MVS, a PatchMatch-based method that integrates SAM-based instance segmentation to better exploits edge information for patch deformation.
Specifically, we first employ the instance segmentation results derived from SAM to adaptively deform the patches for matching cost and propagation, thereby accommodating the distinct characteristics of different pixels. 
Moreover, we employ multi-scale matching cost and propagation scheme to extract diverse information, addressing the challenges posed by textureless areas. 
To optimize memory consumption, we introduce an architecture promoting multi-scale consistency in parallel, consequently reducing the program's runtime.

\par
Moreover, we propose the spherical gradient refinement to optimize previous refinement strategies.
Concerning with normal refinement, we randomly select two orthogonal unit vectors perpendicular to the current normal for perturbation and incorporate gradient descent to further refine perturbation directions in subsequent rounds, thereby improving the accuracy for each hypothesis.
Regarding depth refinement, we adopt pixelwise search interval derived from the deformed patch for local perturbations.

\par
Furthermore, we introduce an EM-based hyperparameter optimization to address the issue of empirical determination of hyperparameters in existing methods. 
By alternately optimizing the aggregated cost and the hyperparameters, we implement an excellent strategy for automatic parameter tuning, thereby facilitating a balanced consideration against diverse information.
Evaluation results on the ETH3D and the Tanks and Temples benchmarks illustrate that our method surpasses the existing state-of-the-art (SOTA) methods. 

In summary, our contributions are as follows:
\begin{itemize}
    \item Based on SAM segmentation, we propose an adaptive patch deformation with multi-scale consistency on both matching cost and propagation to better utilize image edge information and memory cost.
    \item We introduce the spherical gradient refinement, which leverages spherical coordinates and gradient descent on normals and employs pixelwise search interval to constrain depths, thereby enhancing search precision.
    \item We propose the EM-based hyperparameter optimization by adopting the EM algorithm to alternately optimizing the aggregate cost and the hyperparameters.
\end{itemize}

\begin{figure*}
    \centering
    \includegraphics[width=\linewidth]{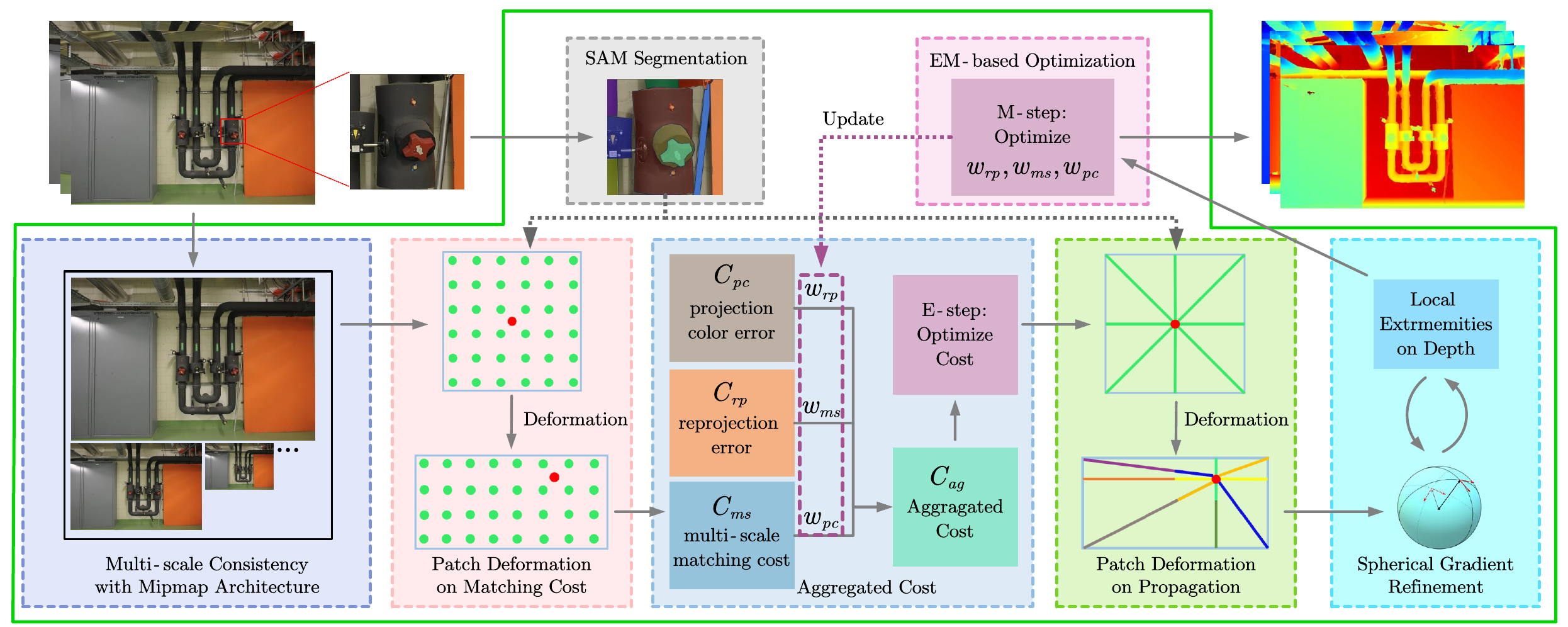}
    \caption{An illustrated pipeline of our proposed method. Images with multi views are initially downsampled and further allocated into our multi-scale architecture. 
    Through leveraging the SAM-based segmentation, we carry out patch deformation on the matching cost to gain multi-scale matching costs $C_{ms}$. By integrating $C_{ms}$ with the projection color error $C_{pc}$ and the reprojection error $C_{rp}$, the aggregated cost is acquired. Then we again employ the SAM-based segmentation for patch deformation in propagation, succeeded by load-balancing within each search domain. Subsequently, we alternately iterates spherical gradient refinement on normals and pixelwise search interval on depths for enhanced accuracy. Finally, we employ EM-based optimization for the hyperparameter tuning of $w_{ms}$, $w_{rp}$, $w_{pc}$ and reassign them for the next iteration procedure. 
    }
    \label{fig:pipeline}
\end{figure*}

\section{Related Work}
\subsubsection{Traditional MVS Methods}
Traditional Multi-View Stereo (MVS) algorithms can primarily be classified into four categories \cite{seitz_comparison_2006}: voxel-based methods \cite{vogiatzis_multiview_2007}, surface evolution-based methods \cite{cremers_multiview_2011} , patch-based methods \cite{bleyer_patchmatch_2011}, and depth-map based methods \cite{yao_recurrent_2019}. Our methodology aligns with the last category, where depth maps are generated from images and their corresponding camera parameters, further leading to point cloud construction via fusion. 
Within this category, PatchMatch-based methods are the most well-known subclass.
Numerous innovative PatchMatch-based methods have been proposed and accomplished a great enhancement in both accuracy and completeness. 
ACMM \cite{xu_multi-scale_2019} uses multi-view consistency and cascading structure to tackle reconstruction of textureless areas, while subsequent works such as ACMMP \cite{xu_multi-scale_2022} further introduce a plane-prior probabilistic graph model and thus provide plane hypothesis for textureless areas.
In contrast, TAPA-MVS \cite{romanoni_tapa-mvs_2019} and PCF-MVS \cite{fink_plane_2019} employ superpixel for image segmentation and planarization of textureless areas. However, the reconstruction performance in textureless areas is contingent upon the actual segmentation and fitting of the superpixels. 
CLD-MVS \cite{li_confidence-based_2020} incorporate a confidence estimator to interpolate unreliable pixels, but their definition way of the confidence makes the result susceptible to occlusion and highlights.
MAR-MVS \cite{xu_marmvs_2020} leverages epipolar geometry to determine the optimal neighborhood images and scale for pixels, yet its fixed patch size limits its adaptability across various application scenarios. 
APD-MVS \cite{yuesong_wang_adaptive_2023} employs patches with adaptive deformation strategy and pyramid architecture, but the time consumption of its iterative process poses a challenge in large-scale datasets.

\subsubsection{Learning-based MVS Methods}
Unlike traditional MVS methods that suffer from hand-crafted image features, learning-based MVS methods typically leverage convolutional neural networks to extract high-dimensional image features, thereby enabling a more rational 3D reconstruction. 
MVSNET \cite{ferrari_mvsnet_2018} has pioneered the construction through introducing differentiable 3D cost volumes using deep neural network, enabling numerous methods for further research.
Certain classic multi-stage methods, including Cas-MVSNet \cite{gu_cascade_2020}, utilize a coarse-to-fine strategy to refine and upscale depth from low-resolution, thereby reducing the cost volumes while expanding the receptive field. 
In terms of memory reduction, several methods like Iter-MVS \cite{wang_itermvs_2022} leverage GRU to regulate the 3D cost volumes along the depth direction. 
Concerning feature extraction, AA-RMVSNET \cite{wei_aa-rmvsnet_2021} aggregates multi-scale variable convolution for adaptive feature extraction. Additionally, MVSTER \cite{avidan_mvster_2022} integrates the transformer architecture into MVS tasks to capture multi-dimensional attention feature information. 
Despite these advancements, it is worth noting that numerous learning-based MVS methods risk severe degradation when applied to target domains that deviate from the training set.

\section{Method}

Given a series of input images $I = \{I_i | i = 1, ..., N \}$, each one with specific camera parameters $P_i = \{K_i, R_i, C_i\}$. Our goal is to estimate the depth map $D_i$ for each image and subsequently merge them into a 3D point cloud. 
Fig. \ref{fig:pipeline} illustrates our overall pipeline, specific design of each component will be detailed in subsequent sections.

\begin{figure}
    \centering
    \includegraphics[width=\linewidth]{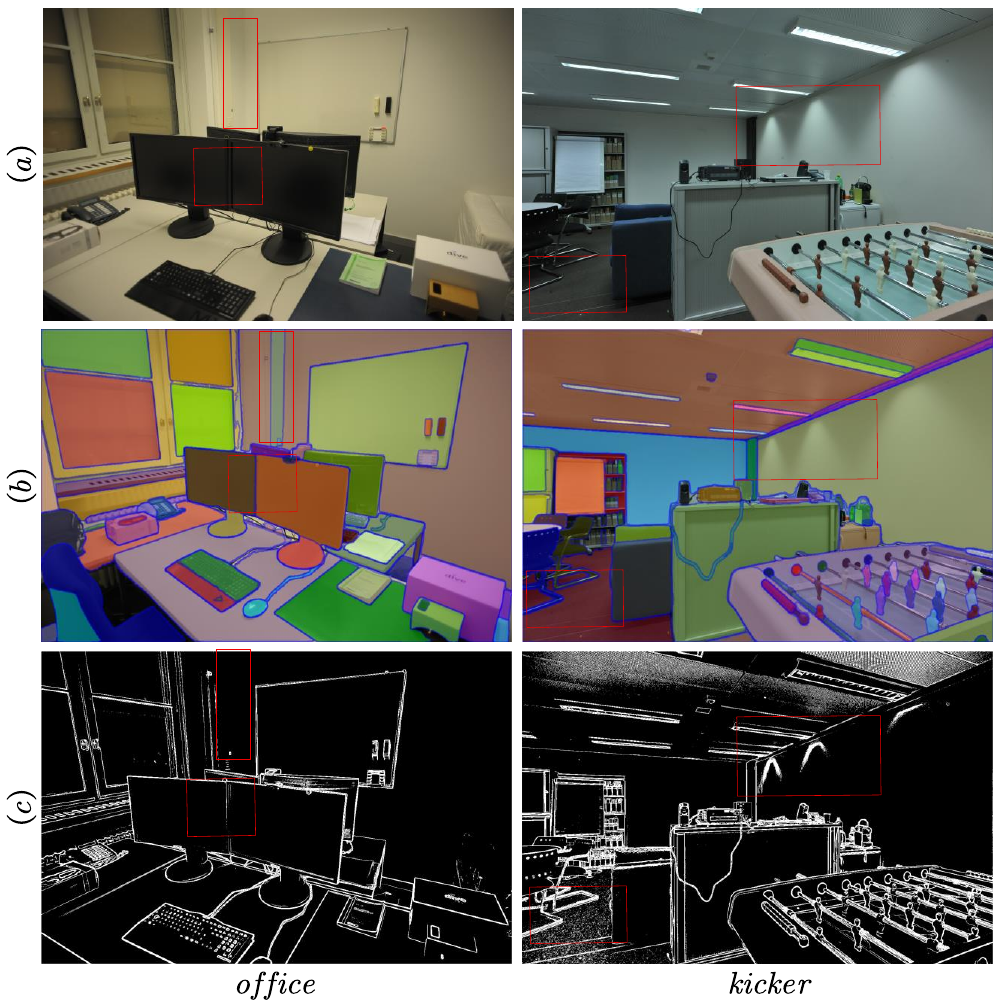}
    \caption{Comparative analysis of patch deformation strategies between the SAM-based instance segmentation and the Canny edge detection on partial scenes of ETH3D dadaset (\emph{office} and \emph{kicker}). From top to bottom, (a), (b) and (c) respectively show the original images, the SAM-based segmentation results and the Canny edge detection results. Representative areas in red boxes illustrate the advantages of SAM-based segmentation over Canny edge detection.}
    
    \label{fig:employing SAM}
\end{figure}

\subsection{Why Using Segment Anything Model?}
The Segment Anything Model (SAM) can effectively discriminate between different instances, extracting subtle edge while neglecting strong illumination disturbances. 
To validate its effectiveness, we conduct the SAM-based instance segmentation and the Canny edge detection for patch deformation on partial scenarios of ETH3D datasets. 

As shown in Fig. \ref{fig:employing SAM}, when confronting with scenarios characterized by extensive similar colors and occlusion like \emph{office}, SAM can effectively separate edges that exhibit similar colors on both sides with inconsistent depths, whereas Canny edge detection simply ignores them. 
Additionally, textureless areas like floors and walls in \emph{kicker} can be effectively separated into different instances through SAM segmentation without illumination interference. In contrast, Canny edge detection incorrectly detects these illumination areas as edges, adversely affecting patch deformation. 

\subsection{Segmentation-Driven Patch Deformation}

\subsubsection{Patch Deformation on Matching Cost}
Some recent methods \cite{wang_patchmatchnet_2021, yuesong_wang_adaptive_2023} attempt to leverage patch deformation to improve matching cost or propagation scheme.
As shown in Fig. \ref{fig:1}, due to their insufficiency in exploiting edge information, they often cross boundary and reference areas with discontinuous depths, thereby yielding unsatisfactory results, especially when confronting with scenarios characterized by extensive similar colors and occlusions like forests and farmlands.
Simultaneously, superpixel-based segmentation approaches \cite{romanoni_tapa-mvs_2019} also struggle in precisely recognizing certain critical edges within these scenarios. They also lack instance semantic information to broaden receptive field, thereby meet pixelwise characteristic.

\begin{figure}
    \centering
    \includegraphics[width=\linewidth]{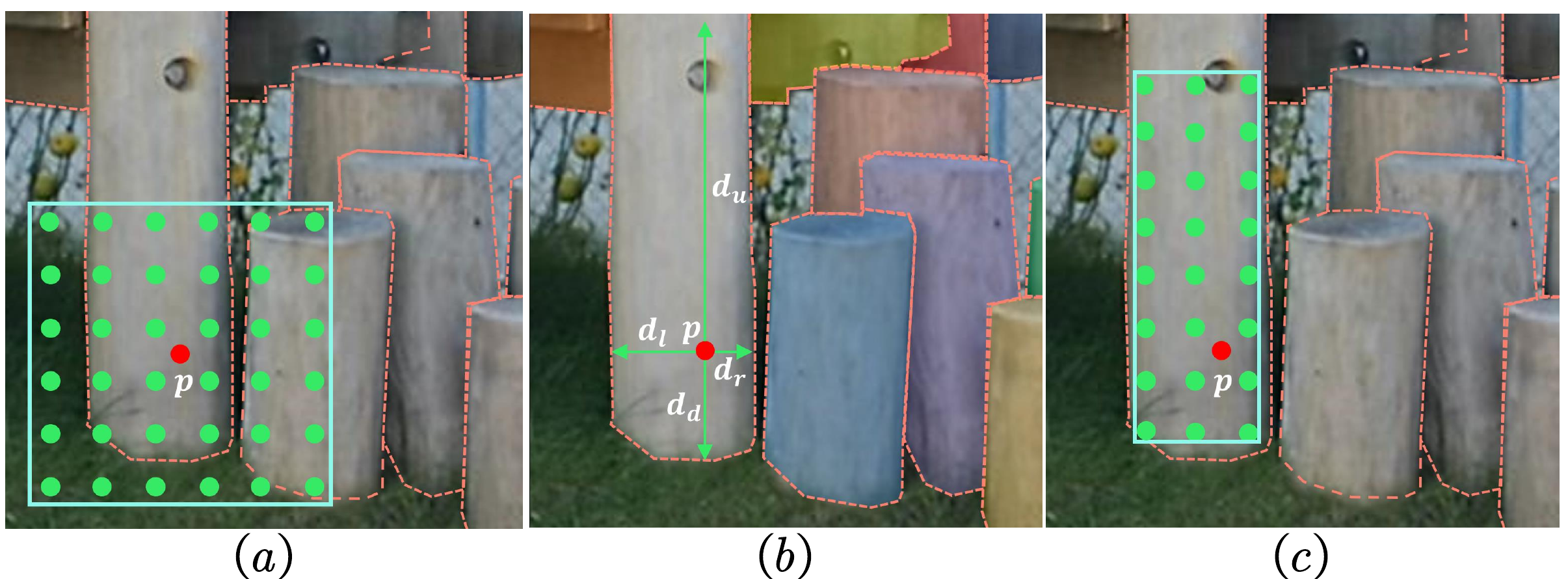}
    \caption{Patch deformation on matching cost. (a) is the matching cost scheme from ACMMP, (b) shows the distance of each directions and (c) illustrates the deformed patch.}
    \label{fig:adaptive_patch}
\end{figure}

SAM segmentation can mitigate this issue as it separates different instances to extract subtle edges information while neglecting robust illumination disturbance. Consequently, we can leverage instance segmentation to better exploit and further introduce edge information into patch deformation.
Specifically, we perform instance segmentation using SAM for input image $I_i$ to generate masks for diverse instances, denoted as $\mathcal{F}$. Hence we have $M = \mathcal{F} (I_i)$, where $M$ is an image mask whose size is consistent with $I_i$. 

For each pixel $p$, we compute the bilateral weighted adaption of normalized cross correlation score (NCC) \cite{leibe_pixelwise_2016} between reference images $I_i$ and source image $I_j$, which can be calculated as follows:
\begin{equation}
\rho\left(p,\mathrm{W}_{p}^{i} \right) =\frac{cov\left( \mathrm{W}_{p}^{i},\mathrm{W}_{p}^{j} \right)}{\sqrt{cov\left( \mathrm{W}_{p}^{i},\mathrm{W}_{p}^{i} \right) cov\left( \mathrm{W}_{p}^{j},\mathrm{W}_{p}^{j} \right)}}
\end{equation}
where $cov$ is weighted covariance, $\mathrm{W}_{p}^{i}$ and $\mathrm{W}_{p}^{j}$ are respectively the corresponding images patches on image $I_i$ and $I_j$.
\par

The goal of minimizing the matching cost is to obtain the optimal matching depths via the computation of color differences. 
However, when objects with varying depths exhibit similar colors, they are susceptible to generating matching inaccuracies, as shown in Fig. \ref{fig:adaptive_patch}(a).
Therefore, we introduce patch deformation to compute matching cost upon the sample patch $\mathrm{W}$ intersecting with different instances.

Specifically, we first measure the distances from the corresponding central pixel $p$ to the left, right, lower and upper boundaries of $M$, denoted respectively as $d_{l}$, $d_{r}$, $d_{d}$, and $d_{u}$.
Then we can deform the shape of $\mathrm{W}$ to match these boundaries. The new shape of deformed patch can be defined as:
\begin{equation}
    \left[ \frac{d_l+d_r}{d_l+d_r+d_d+d_u}L, \frac{d_d+d_u}{d_l+d_r+d_d+d_u}L \right] 
\end{equation}
where $L$ denotes the side length of the square patch before patch deformation.
Additionally, we reposition the patch's center by adding an offset:
\begin{equation}
    \Delta o(p)=\left( \frac{d_l-d_r}{d_l+d_r}L_h, \frac{d_d-d_u}{d_u+d_d}L_v\right) 
\end{equation}
where $L_h$ and $L_v$ are respectively the horizontal and vertical length of deformed patch.
The new center of the sample patch now becomes $p+\Delta o(p)$.

Both patch deformation and center offset allow pixels positioned at boundary regions to orient their patches more intensively towards the center of its own instance.
Enhancing the receptive field for homogenous pixels in such approach can yield more robust results, consequently reducing potential errors in estimation.
Note that considering the runtime, we restrict the number of calculations for each window such that the number of calculations after deformation never surpasses the initial total number $(L/ 2)^2$.

\subsubsection{Patch Deformation on Propagation}
After SAM-based instance segmentation, pixels within the same instance typically exhibit similar depths, whereas noticeable depth discontinuities frequently arise at the boundaries between instances. Considering that propagation involves updating potential depths and normals within the surrounding area for each pixel, depth discontinuities will inevitably impact propagation. Consequently, we leverage patch deformation to adaptively alter the propagation scheme.

\begin{figure}
    \centering
    \includegraphics[width=\linewidth]{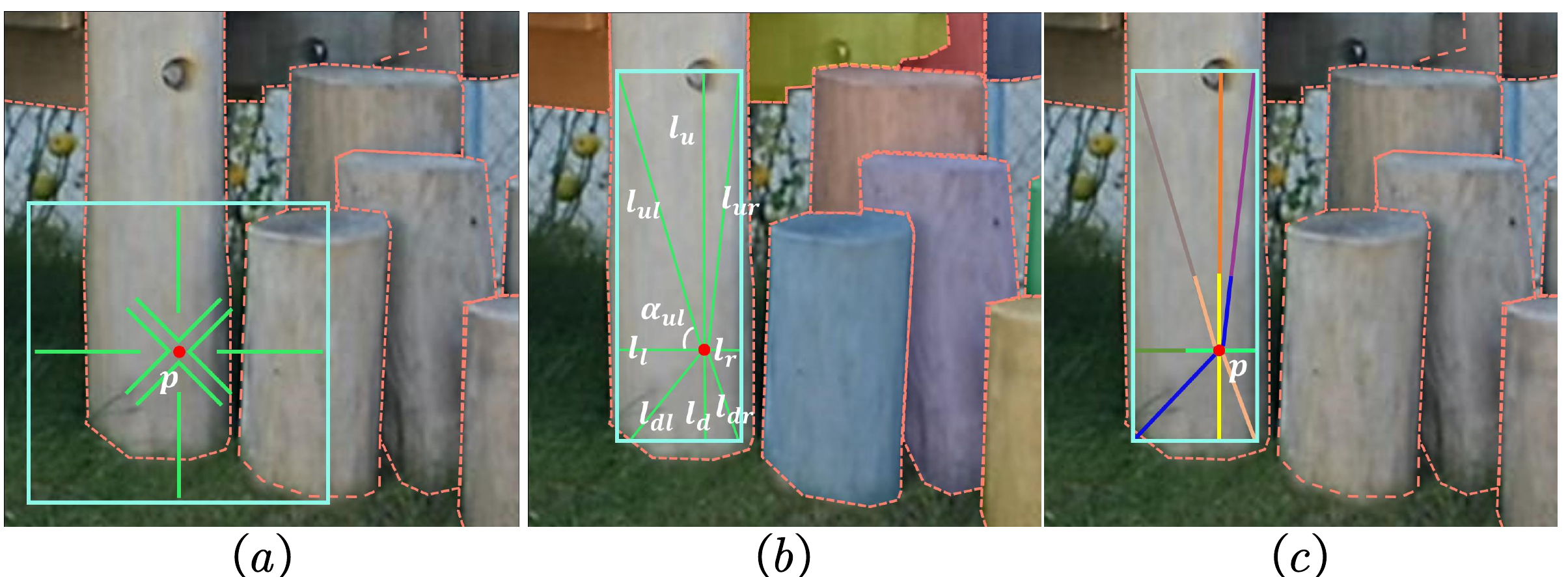}
    \caption{Patch deformation on propagation. (a) is the propagation pattern of ACMMP, (b) depicts the length of each propagation branch, and (c) illustrates different search domains with different colors.}
    \label{fig:propagation}
\end{figure}

\par
The adaptive checkerboard propagation scheme \cite{xu_multi-scale_2019} is conducted by introducing the optimal hypotheses from four near and four far search domains, as illustrated in Fig. \ref{fig:propagation} (a). 
However, his search domain between two adjacent diagonal directions is too dense, which leads to an imbalanced search space density and a risk of selecting redundant values. Hence we modify its oblique direction into a straight line extending to the corner of each patch.

\par
Subsequently, we propose patch deformation on propagation via SAM, which adjusts the propagation patch shape and direction for each pixel. As illustrated in Fig. \ref{fig:propagation} (b), we adapt the propagation directions according to the shape of the surrounding mask. 
Specifically, denoting $l_l$, $l_r$, $l_d$, and $l_u$ as the length from the central pixel $p$ to the left, right, lower and upper edges of the patch, respectively, we obtain:
\begin{equation}
    l_u=\frac{d_u}{d_u+d_d}L_v, l_l=\frac{d_l}{d_l+d_r}L_h
\end{equation}

Both $l_r$ and $l_d$ can be obtained similarly. Therefore, the directions and lengths of slanted branch $l_{ul}$ is given by:
\begin{equation}
    l_{ul}=\sqrt{{l_u}^2+{l_l}^2}, \alpha _{ur}=\mathrm{arc}\tan \left( \frac{l_{u}}{l_l} \right) 
\end{equation}
where $l_{ul}$ refers to the length of the up-right branch, and $\alpha _{ur}$ represents the angle between the upward branch and the up-right branch. Corresponding lengths and directions of other three slanted branches can be obtained similarly.

\par

Having adjusted all directions and lengths, we encounter another challenge: the searching domain for each branch is unbalanced. Since the process of selecting a pixel with the minimal cost is essentially a spatial neighborhood search, an imbalance will emerge due to the different length of branches. The search along a shorter branch is suffered from unreliable results due to its minor search domain.
\par
To address this, we accordingly modify the searching strategy in the propagation scheme, as shown in Fig. \ref{fig:propagation} (c).
Specifically, we employ eight different colors to depict separate search domains on the eight directions centered on $p$. 
Instead of taking the central pixel $p$ as the dividing point, we use the midpoint of the sum of the lengths of two opposite branches to divide the search domain.
In experiments, pixels with the same color are grouped into the same domain, with CUDA operators balance the load of searching for minima within each color-specific region.
Therefore, our proposed strategy ensures load-balance across all directions and allows for faster convergence.

\begin{figure}
    \centering
    \includegraphics[width=0.95\linewidth]{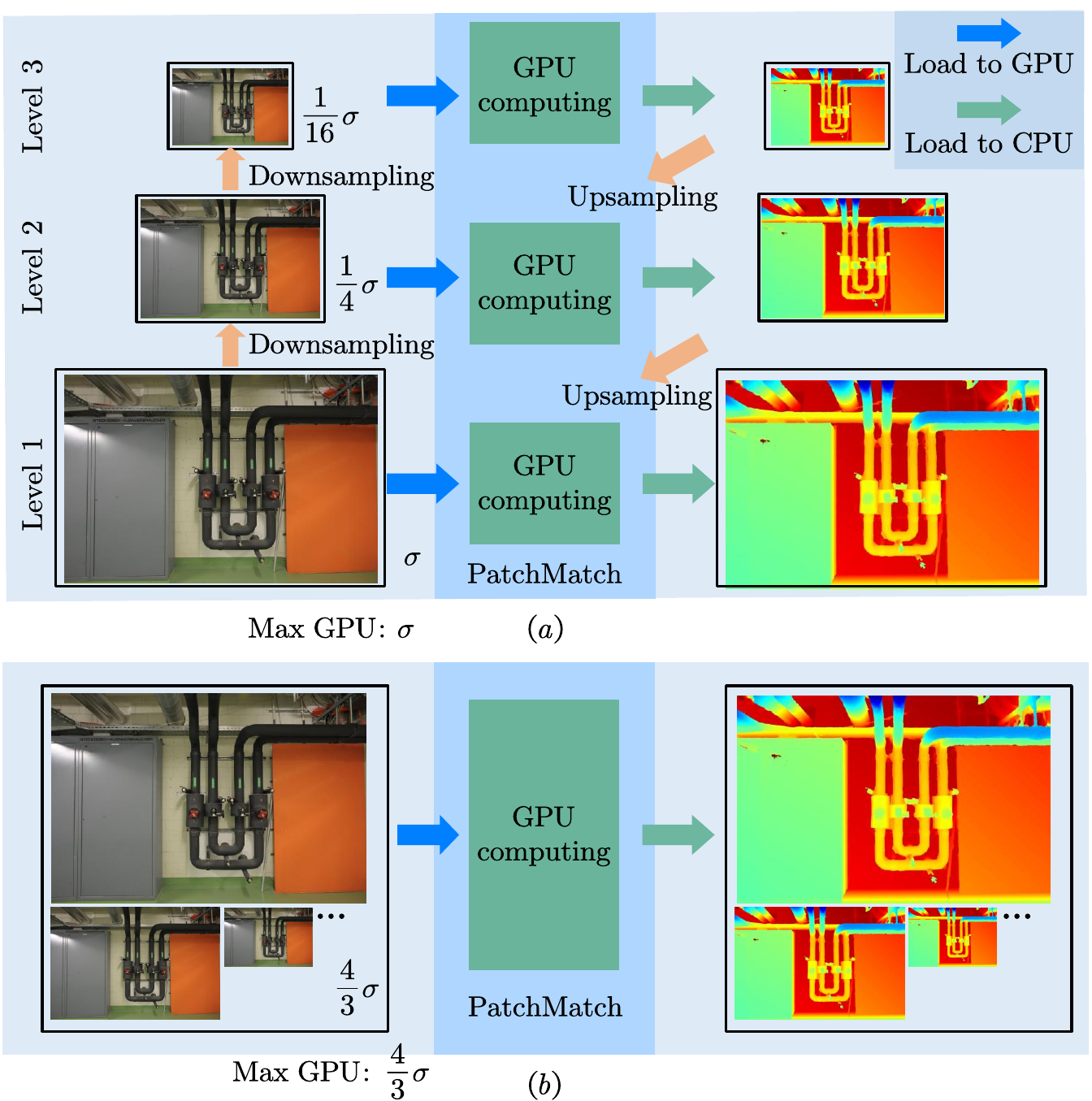}
    \caption{Different design architectures between ACMMP and our method. (a) illustrates the cascading network architectures employed in ACMMP, whereas (b) depicts our method with multi-scale architecture.}
    \label{fig:multi-scale}
\end{figure}

\subsubsection{Multi-scale Consistency}

Many conventional methods adopt cascading architectures by sequentially loading different scales of images into GPU, as shown in Fig. \ref{fig:multi-scale} (a). This may result in a time-consuming performance due to the limited transfer speed between CPU and GPU. Therefore, we draw inspiration from mipmap \cite{williamsPyramidalParametrics1983} in computer graphics, a technique to load different scales of images in parallel at once, to replace the previous cascading architecture into our proposed parallel architecture.

Specifically, we first perform image downsampling in the CPU. Subsequently, multi-scale images are assembled and loaded together into the GPU, as depicted in Fig. \ref{fig:multi-scale} (b). Then multi-scale images are processed together through matching cost, propagation and refinement in the GPU. Finally, all predicted depth images are transferred back into the CPU. 
Denoting the maximum memory consumption of ACMMP cascading architectures as $\sigma$, and the number of memory read operations as $k$, this technique enables us to load all scales of images in the GPU memory at a reasonable cost of $\frac{4}{3}\sigma$ instead of sequentially loading images, thereby eliminating the need for $k-1$ additional memory read operations. 

Based on this architerture, we further introduce multi-scale consistency on matching cost and propagation. Regarding matching cost, we first apply SAM segmentation on the $k$-th level downsampled image. Based on segmentation results, we construct deformed patch and further compute $k$-th level matching cost, denoted as $c_k$.
Therefore, the multi-scale matching cost is given by:
\begin{equation}
    C_{ms} = \frac{\sum_k{c_k}}{k} \,\
\end{equation}

\par
Concerning with propagation, the multi-scale consistency aggregates the search domain for all scales in each direction, yielding a total of eight distinct search domains.  Conclusively, eight values with the lowest cost within each domain are chose as new hypothesis for further computation.

\subsubsection{Aggregated Cost}

During the patch-matching phase, we consider not only the multi-scale matching cost $C_{ms}$, but also the reprojection error $C_{rp}$ and the projection color gradient error $C_{pc}$. $C_{rp}$ proposed in ACMMP validates depth estimation from geometric consistency.
$C_{pc}$ measures color consistency between current pixel $p_i$ in reference image $I_i$ and its corresponding pixel $p_j$ in source images $I_j$:
\begin{equation}
    C_{pc}=\max \left\{ \left\| \nabla I_j\left( p_j \right) -\nabla I_i\left( p_i \right) \right\| ,\tau \right\} 
\end{equation}
where $\nabla$ represents the Laplacian Operator, $p_j$ denotes pixel in image $I_j$ the projected by pixel $p_i$ in  $I_i$, and $\tau$ is the truncation threshold to robustify the cost against outliers. With these terms, our the aggregated costs $C_{ag}$ can be given by:
\begin{equation}
    C_{ag}=w_{ms}C_{ms}+w_{rp}C_{rp}+w_{pc}C_{pc}
\end{equation}
where $w_{ms}$, $w_{rp}$, and $w_{pc}$ respectively represent the aggregation weights of each component.

\par

\subsection{Spherical Gradient Refinement}
Two types of refinement strategies are adopted in ACMMP: 1. \textbf{Local perturbations}, which is the local search conduct by perturbing the current depth and normal with a small value; 2. \textbf{Random selection}, which achieves global search to suit potential depth discontinuities by assigning a random value. Since the edge information has already been segmented out through SAM, we only need to consider local perturbations. 
Given depth $d$ and normal $n = (n_x, n_y, n_z)$ in Cartesian coordinates, new depth $d'$ and normal $n'$ after the local perturbation can be defined by:
\begin{equation}
\begin{cases}
    d'\gets d+\delta_d \\
    n'\gets \mathbf{VN}\left( n_x+\delta_x ,n_y+\delta_y ,n_z+\delta_z \right) \\
\end{cases}
\end{equation}
where $\mathbf{VN}$ is a normalization function ensuring $\|n'\|=1$, and $\delta$ denotes a random value chosen from a fixed interval.

However, the strategy is incompatible with the definition of normal. It introduces a higher sensitivity to axes with smaller values during the search process, resulting in an unequal ratio of change on $xyz$ axes. Therefore, we propose the spherical gradient descent refinement, which utilize a structured representation to converge more accurate hypotheses.

\subsubsection{Spherical Coordinate}
\begin{figure}
    \centering
    \includegraphics[width=1\linewidth]{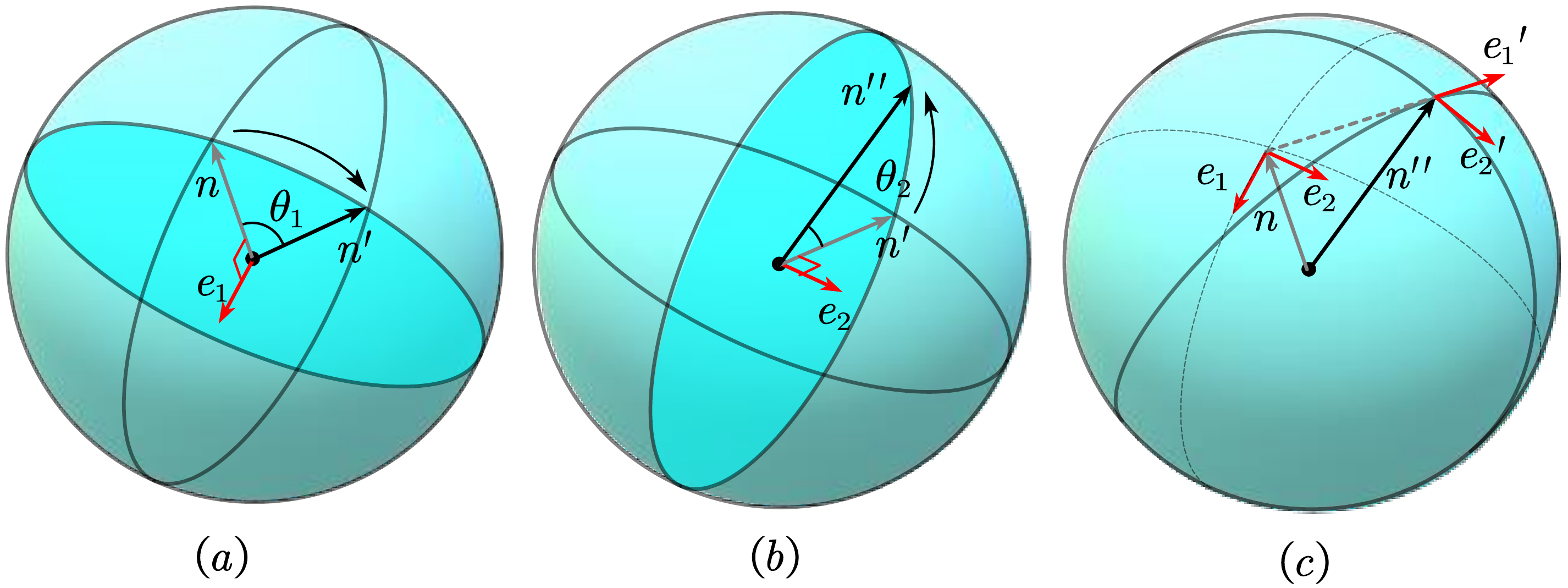}
    \caption{Spherical Gradient Refinement Procedure. 
    (a) illustrates the rotation from $n$ to $n'$, (b) illustrates the rotation from $n'$ to $n''$. (c) respectively indicates two old and new orthogonal perturbation directions $e_1$, $e_2$ and $e_1'$, $e_2'$.
    }
    \label{fig:refine}
\end{figure}

As shown in Fig. \ref{fig:refine}, given the normalized normal, we first randomly choose two orthogonal vectors, $e_1$ and $e_2$, perpendicular to the normal $n$ as the perturbation direction. We then use the angles $\theta_1$ and $\theta_2$ as the degree of rotation for iterative refinement.  The normal first undergoes a counterclockwise rotation by $\theta_1$ degrees around $e_1$ as the rotation axis. Subsequently, the normal is further rotated counterclockwise by $\theta_2$ degrees around $e_2$ as the rotation axis. According to Rodrigues' rotation formula, the ultimate updated normal $n''$ is given by:
\begin{equation}
\begin{cases}
n' = cos\theta_1 \cdot n + sin\theta_1 (e_1 \times n)\\
n'' = cos\theta_2 \cdot n' + sin\theta_2 (e_2 \times n')
\end{cases}
\end{equation}
This is analogous to sliding a vertex directed by the normal on the surface of a sphere, which ensures the preservation of normalization for the normal vector both before and after rotation. By finding two orthogonal bases perpendicular to the normal for refinement, it can be ensured that perturbations in each direction are equivalent. This approach aligns more closely with the geometric essence of the normal, which is defined on a sphere rather than individual axes in the $xyz$ coordinate system. As a result, our approach boosts the robustness and stability during the refinement process.

\begin{figure*}
\centering
\includegraphics[width=\linewidth]{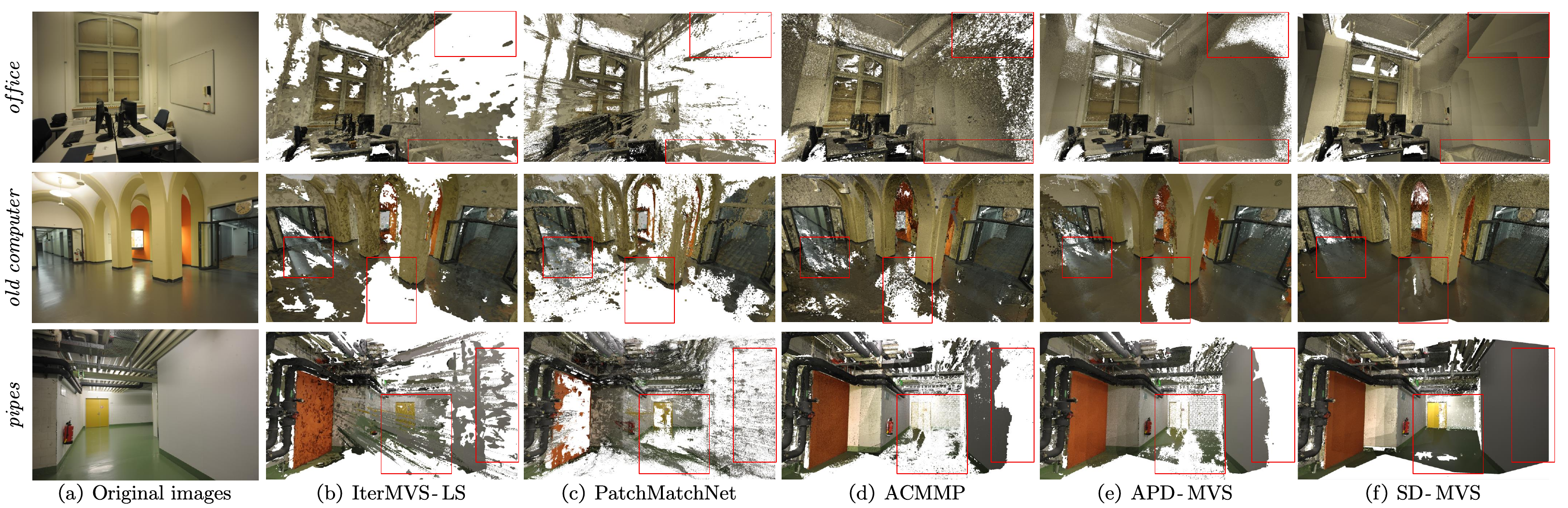}
\caption{An illustration of the qualitative results on partial scenes of ETH3D datasets (\emph{office}, \emph{old computer}, and \emph{pipes}). Some challenging areas are shown in red boxes. It is obvious that our methods outperform others, especially in large textureless areas.}
\label{fig: eth3d results}
\end{figure*}

\subsubsection{Gradient Descent}
We also utilize gradient descent in our method. The primary merit of gradient descent lies in its ability to logically restrict the search space to the vicinity of probable solutions.
Denoting the number of total iterations as $N_{max}$, the rotation angle $\theta$ for the $i$-th round is randomly selected from range $\left[0, 5 * 2^{N_{max}-i}\right]$.  After one round of refinement for depth $d$ and normal $n$, we determine the new direction for local perturbations $e_1'$ and $e_2'$ based on the result of the previous search. As such, we get:
\begin{equation}
\begin{cases}
    e_1'\gets n'' - n\\
    e_2'\gets e_1'\times n''
\end{cases}
\end{equation}
Here, $e_1'$ is aligned with the vector sum of the previous round's perturbation, while $e_2'$ is a vector perpendicular to both $n'$ and $e_1'$, as shown in Fig. \ref{fig:refine}(c). 
The primary merit of gradient descent lies in its ability to restrict the search domain of neighbourhood solutions. Each round of search takes place on the orthogonal plane defined by the previous search direction and the current normal direction, thereby enabling faster convergence to the optimal solution.

\subsubsection{Pixelwise Depth Interval Search}
ACMMP employs a fixed interval for local perturbations on depth, while static perturbation range cannot adapt well to locally varying scene depth.
Addressing this, we introduce pixelwise depth search interval chosen within the deformed patch.

Specifically, for each pixel, we extract the depth values of all pixels encompassed by its deformed patch, and choose the maximal and minimal values from this set as depth boundary for perturbations. Additionally, considering our iterative refinement strategy, during the $i$-th iteration, the pixelwise search interval is chosen within the deformed patch gained from $i$-th downsampled image, thereby narrowing the perturbation interval to yield more accurate hypothesis.

\subsection{EM-based Hyperparameters Optimization}

While computing the aggregated matching cost, the hyperparameters of each component is typically determined empirically, which may result in suboptimal outcomes for different scenes. To mitigate this, we leverage the Expectation-Maximization (EM) algorithm to alternately optimize the hyperparameters and the aggregated cost, thereby enhancing both the robustness and effectiveness of our method.

\subsubsection{E-Step: Optimize $C_{ag}$}

By fixing $w_{ms}$, $w_{rp}$, and $w_{pc}$, we can optimize the aggregated cost $C_{ag}$, formulated as:
\begin{equation}
   \underset{C_{ms},C_{rp},C_{pc}}{\min} C_{ag}=w_{ms}C_{ms}+w_{rp}C_{rp}+w_{pc}C_{pc}
\end{equation}
After optimization, we can get the optimal depth estimation under current hyperparameters.

\subsubsection{M-Step: Optimize $w_{ms}$, $w_{rp}$, $w_{pc}$} 
By fixing $C_{ms}$,$C_{rp}$ and $C_{pc}$, we can optimize $w_{ms}$, $w_{rp}$ and $w_{pc}$, defined by:
\begin{equation}
\begin{aligned}
\underset{w_{ms},w_{ms},w_{pc}}{\min} & C_{ag}=w_{ms}C_{ms}+w_{rp}C_{rp}+w_{pc}C_{pc}, \\
s.t. \quad
        &w_{ms}+w_{rp}+w_{pc}=1, \\
	&w_{ms}, w_{rp}, w_{pc} > \eta\\
\end{aligned}
\end{equation}
All hyperparameters are required to exceed a minimal value $\eta$, and we implement a normalization constraint ensuring that their sum equals $1$ to mitigate significant variances.
Following the E-step optimization, we can alternatively optimize the hyperparameters and feed them back into the E-step for the next round of aggregated cost optimization.
\par
Since it may be challenging to obtain the analytical solution to the optimization problem in M-step, we will use numerical optimization methods such as Newton's method \cite{newton} to obtain the optimal solutions for $w_{ms}$, $w_{rp}$, and $w_{pc}$. A comprehensive formula derivation of the optimization can be found in supplementary material.
\par
In practical situations, there might be partial pixels with depth estimation errors when all pixels are selected. Hence, we only select pixels where SIFT features can be matched between different images, and then calculate the aggregate cost between the pixels corresponding to these features. 

\begin{table}
  \centering
  \renewcommand{\arraystretch}{1.05} 
    \resizebox{\linewidth}{!}{
        \begin{tabular}{c|ccc|ccc}
        \hline
        \multirow{2}{*}{Method} & \multicolumn{3}{c|}{Train} & \multicolumn{3}{c}{Test} \\
        \cline{2-7} & Acc. & Comp. & F$_1$ & Acc. & Comp. & F$_1$ \\   
        \hline 
        \multirow{1}{*}{PatchMatchNet} & 64.81 & 65.43 & 64.21 & 69.71 & 77.46 & 73.12 \\  
        \multirow{1}{*}{IterMVS-LS} & 79.79 & 66.08 & 71.69 & 84.73 & 76.49 & 80.06 \\  
        \multirow{1}{*}{MVSTER} & 68.08 & 76.92 & 72.06 & 77.09 & 82.47 & 79.01 \\  
        \multirow{1}{*}{EPP-MVSNet} & 82.76 & 67.58 & 74.00 & 85.47 & 81.79 & 83.40 \\  
        \multirow{1}{*}{EPNet} & 79.36 & 79.28 & 79.08 & 80.37 & \textbf{87.84} & 83.72 \\  
        \hline 
        \multirow{1}{*}{COLMAP} & \textbf{91.85} & 55.13 & 67.66 & \textbf{91.97} & 62.98 & 73.01 \\  
        \multirow{1}{*}{PCF-MVS} & 84.11 & 75.73 & 79.42 & 82.15 & 79.29 & 80.38 \\  
        \multirow{1}{*}{MAR-MVS} & 81.98 & 77.19 & 79.21 & 80.24 & 84.18 & 81.84 \\  
        \multirow{1}{*}{ACMP} & 90.12 & 72.15 & 79.79 & 90.54 & 75.58 & 81.51 \\  
        \multirow{1}{*}{ACMMP} & \underline{90.63} & 77.61 & 83.42 & \underline{91.91} & 81.49 & 85.89 \\  
        \multirow{1}{*}{APD-MVS} & 89.14 & \textbf{84.83} & \underline{86.84} & 89.54 & 85.93 & \underline{87.44} \\  
        \hline 
        \multirow{1}{*}{SD-MVS (ours)} & 89.63 & \underline{84.52} & \textbf{86.94} & 88.96 & \underline{87.49} & \textbf{88.06} \\  
        
        \hline
        \end{tabular}%
    }
  \caption{Quantitative results on ETH3D benchmark at threshold $2cm$ . Our method accomplishes the best F$_1$ score.}
  \label{table: eth3d results}%
\end{table}%

\section{Experiments}

\subsection{Datasets and Implementation Details}
We evaluate our work on both ETH3D high-resolution benchmark \cite{schoeps2017cvpr} and Tanks and Temples benchmark (TNT) \cite{Knapitsch2017}. We compare our work against state-of-the-art learning-based methods including PatchMatchNet \cite{wang_patchmatchnet_2021}, IterMVS-LS \cite{wang_itermvs_2022}, MVSTER \cite{avidan_mvster_2022}, EPP-MVSNet \cite{ma_epp-mvsnet_2021}, EPNet \cite{EPNet} and traditional MVS methods including COLMAP \cite{leibe_pixelwise_2016}, PCF-MVS \cite{fink_plane_2019}, MAR-MVS \cite{xu_marmvs_2020}, ACMP \cite{xu_planar_2020}, ACMMP \cite{xu_multi-scale_2022} and APD-MVS \cite{yuesong_wang_adaptive_2023}. 

Note that experiments is carried out on downsampled images with half of the original resolution in ETH3D, and on original images in TNT. Concerning parameter setting, $\{w_{ms}, w_{rp}, w_{pc}, L, k, \tau, N_{max}, \eta \} = \{1, 0.2, 0.2, 11, 3, 2, 3, 0.1\}$. In cost calculation, we take the matching strategy of every other row and column. 

Our method is implemented on a system equipped with an Intel(R) Core(TM) i7-10700 CPU @ 2.90GHz and an NVIDIA GeForce RTX 3080 graphics card. We take ACMP \cite{xu_planar_2020} as the backbone of our method. 

\begin{table}
    \centering
    \renewcommand{\arraystretch}{1.05} 
    \resizebox{\linewidth}{!}{
        \begin{tabular}{c|ccc|ccc}
        \hline
        \multirow{2}{*}{Method} & \multicolumn{3}{c|}{Intermediate} & \multicolumn{3}{c}{Advanced} \\
        \cline{2-7} & Pre. & Rec. & F$_1$ & Pre. & Rec. & F$_1$ \\   
        \hline 
        \multirow{1}{*}{PatchMatchNet} & 43.64 & 69.37 & 53.15 & 27.27 & 41.66 & 32.31 \\  
        \multirow{1}{*}{CasMVSNet} & 47.62 & 74.01 & 56.84 & 29.68 & 35.24 & 31.12 \\  
        \multirow{1}{*}{IterMVS-LS} & 47.53 & 74.69 & 56.94 & 28.70 & 44.19 & 34.17 \\  
        \multirow{1}{*}{MVSTER} & 50.17 & \underline{77.50} & 60.92 & 33.23 & 45.90 & 37.53 \\  
        \multirow{1}{*}{EPP-MVSNet} & 53.09 & 75.58 & 61.68 & \textbf{40.09} & 34.63 & 35.72 \\
        \multirow{1}{*}{EPNet} & \textbf{57.01} & 72.57 &  \textbf{63.68} & 34.26 & \textbf{50.54} & \textbf{40.52} \\ 
        \hline 
        \multirow{1}{*}{COLMAP} & 43.16 & 44.48 & 42.14 & 31.57 & 23.96 & 27.24 \\  
        \multirow{1}{*}{PCF-MVS} & 49.82 & 65.68 & 55.88 & 34.52 & 35.36 & 35.69 \\  
        \multirow{1}{*}{ACMP} & 49.06 & 73.58 & 58.41 & 34.57 & 42.48 & 37.44 \\  
        \multirow{1}{*}{ACMMP} & 53.28 & 68.50 & 59.38 & 33.79 & 44.64 & 37.84 \\  
        \multirow{1}{*}{APD-MVS} & \underline{55.58} & 75.06 & \underline{63.64} & 33.77 & \underline{49.41} & 39.91 \\  
        \hline 
        \multirow{1}{*}{SD-MVS (ours)} & 53.78 & \textbf{77.63} & 63.31 & \underline{35.53} & 47.37 & \underline{40.18} \\  
        
        \hline
        \end{tabular}%
    }
    \caption{Quantitative results on TNT dataset. Our method accomplishes competitive F$_1$ score with SOTA methods.}
    \label{table: TNT results}%
\end{table}%

\subsection{Results on ETH3D and TNT}
Qualitative results on ETH3D are illustrated in Fig. \ref{fig: eth3d results}. It is obvious that our method reconstructs the most comprehensive results, especially in large textureless areas like floors, walls and doors, without introducing conspicuous detail distortion. More qualitative results on ETH3D and TNT benchmark can be referred in supplementary material. 

\par
Tab. \ref{table: eth3d results} and Tab. \ref{table: TNT results} respectively present quantitative results on the ETH3D and the TNT benchmark. Note that the first group is learning-based methods and the second is traditional methods. Meanwhile, the best results are marked in bold while the second-best results are underlined.
Our method achieves the highest F$_1$ score on ETH3D datasets, giving rise to state-of-the-art performance. Meanwhile, our method achieves competitive results with SOTA methods in TNT datasets like EPNET and APD-MVS, falling short by less than $0.5\%$ in F$_1$ score. Especially, our method shows significant improvement in completeness in both datasets, demonstrating its robustness in recovering textureless areas. 

\begin{table}
    \centering
    \renewcommand{\arraystretch}{1.05} 
    \resizebox{\linewidth}{!}{
        \begin{tabular}{c|ccc|ccc}
        \hline
        \multirow{2}{*}{Method} & \multicolumn{3}{c|}{$2cm$} & \multicolumn{3}{c}{$10cm$} \\
        \cline{2-7} & Acc. & Comp. & F$_1$ & Acc. & Comp. & F$_1$ \\   
        \hline  
        \multirow{1}{*}{w/. ACM. Cost} & \textbf{90.16} & 74.61 & 81.27 & \textbf{98.01} & 89.04 & 93.16 \\  
        \multirow{1}{*}{w/o. Adp. Cost} & 89.92 & 78.01 & 83.42 & 97.92 & 91.87 & 94.71  \\  
        \multirow{1}{*}{w/o. Mul. Cost} & 89.84 & 79.94 & 84.55 & 97.9 & 93.36 & 95.53  \\  
        \hline
        \multirow{1}{*}{w/. ACM. Pro.} & 89.83 & 79.96 & 84.52 & 97.91 & 93.58 & 95.54  \\  
        \multirow{1}{*}{w/o. Adp. Pro.} & 89.57 & 81.74 & 85.38 & 97.81 & 94.96 & 96.29  \\  
        \multirow{1}{*}{w/o. Mul. Pro.} & 89.69 & 81.97 & 85.54 & 97.87 & 95.17 & 96.44  \\  
        \hline
        \multirow{1}{*}{w/o. Ref.} & 86.75 & 70.45 & 77.6 & 97.04 & 85.37 & 90.72  \\  
        \multirow{1}{*}{w/. Gip. Ref.} & 89.3 & 78.51 & 83.43 & 97.74 & 91.56 & 94.48  \\  
        \multirow{1}{*}{w/. ACM. Ref.} & 89.42 & 79.83 & 84.25 & 97.79 & 92.64 & 95.11  \\  
        \hline
        \multirow{1}{*}{w/o. EM A} & 89.74 & 78.16 & 83.45 & 97.89 & 91.78 & 94.57 \\  
        \multirow{1}{*}{w/o. EM B} & 89.45 & 79.87 & 84.27 & 97.81 & 93.05 & 95.3  \\
        \hline
        \multirow{1}{*}{SD-MVS} & 89.63 & \textbf{84.52} & \textbf{86.94} & 97.85 & \textbf{96.74} &  \textbf{97.28} \\  
        \hline
        \end{tabular}%
    }
    \caption{Quantitative results of the ablation studies on ETH3D benchmark to validate each proposed component.}
    \label{table: ablation study}%
\end{table}%

\subsection{Memory and Runtime Comparison}

\begin{figure}
    \centering
    \includegraphics[width=\linewidth]{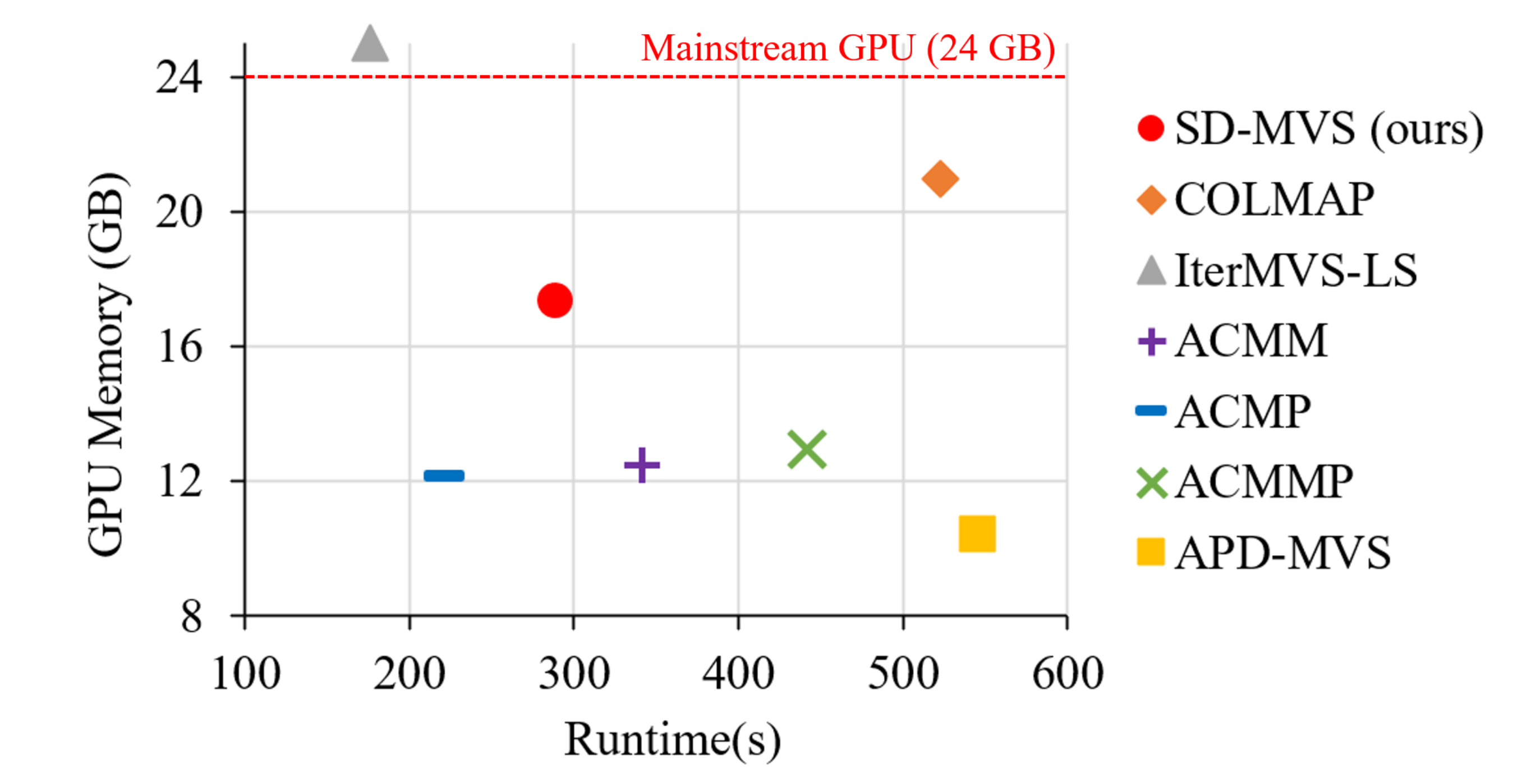}
    \caption{GPU memory usage (GB) and runtime (second) between different methods on ETH3D training datasets.}
    \label{fig:mem and time}
\end{figure}

To demonstrate the efficiency of our method, we compare both GPU memory usage and runtime among various methods on ETH3D training datasets, as depicted in Fig. \ref{fig:mem and time}. Note that all experiments are executed on original images whose number have been standardized to $10$ across all scenes. Moreover, to exclude the impact of unrelated variables, all methods are conducted on a same system, whose hardware configuration has been specified in previous section.

\par
Concerning learning-based methods, while IterMVS-LS exhibits the shortest runtime, its memory overhead exceeds the maximum capacity of mainstream GPUs. Other state-of-the-art (SOTA) learning-based methods also suffer from excessive memory consumption, making them impractical for the reconstruction of large-scale outdoor scenarios. 

Although SD-MVS consumes approximately one-third more memory usage than traditional SOTA methods like APD-MVS and ACMMP, our runtime is only half of them, thanks to our multi-scale consistency architecture. Therefore, our method strikes the optimal balance between time and memory usage without sacrificing performance, demonstrating its effectiveness and practicality.

\par

\subsection{Ablation Studies}

We validate the rationale behind the design of each part of our method through ablation studies, as shown in Tab. \ref{table: ablation study}.

\subsubsection{Matching Cost with Adaptive Patch} 
In terms of matching cost, we respectively remove patch deformation (w/o. Adp. Cost), multi-scale consistency (w/o. Mul. Cost) and both of them (w/. ACM. Cost). 
Since w/. ACM. Cost has neither deformable nor multi-scale, it produces the worst results. w/o. Mul. Cost slightly outperformed w/o. Adp. Cost, yet both are inferior to SD-MVS, implying that patch deformation contribute more than multi-scale consistency.

\subsubsection{Adaptive Propagation with Load-balancing} 
In terms of propagation, we respectively remove patch deformation (w/o. Adp. Pro.), multi-scale consistency (w/o. Mul. Pro.) and apply propagation scheme from ACMMP (w/. ACM. Pro.).
Given that patches in ACMMP do not deform in accordance with the patch, its performance fell short of expectations. Both w/o. Adp. Pro. and w/o. Mul. Pro. delivered similar results, yet fell short in comparison to SD-MVS, indicating that both patch deformation and multi-scale consistency on propagation are equally crucial.

\subsubsection{Spherical Gradient Refinement} 
In terms of refinement, we respectively remove refinement (w/o. Ref.), exchange the refinement module into Gipuma \cite{galliani_massively_2015} (w/. Gip. Ref.) and switch the refinement module into ACMMP (w/. ACM. Ref.). 
As observed, the absence of refinement significantly diminishes the results. However, introducing Gipuma refinement brings about noticeable progress, with further advancements achieved after adopting ACMMP refinement. Nonetheless, both refinement methods are worse than SD-MVS, proving the necessity of spherical gradient refinement.

\subsubsection{EM-based Hyperparameters Optimization}
We conduct two experiments (w/o. EM A and w/o. EM B) by removing EM-based Optimization and respectively setting $(w_{ms}, w_{rp}, w_{pc})$ to $(1,0.5,0.5)$ and $(1,0.2,0.2)$. The results highlight the impact of hyperparameter settings on the final results.
 Furthermore, their inferior performances compared to SD-MVS evidences the importance of automatic parameter tuning by the proposed EM-based Optimization.

\section{Conclusion}

In this paper, we presented SD-MVS, a novel MVS method designed to effectively address challenges posed by textureless areas. 
The proposed method consists of an adaptive patch deformation with multi-scale consistency, a spherical gradient refinement and EM-based hyperparameter optimization. 
Our method has achieved state-of-the-art performance on ETH3D high-resolution benchmark, while being memory-friendly and with less time cost. 
In the future, we will tackle difficulty in highlight areas in matching cost and view selection strategy in pursuit of superior performance.

\section{Acknowledgements}

This work was supported by the National Natural Science Foundation of China under Grant 62172392, the Central Public-interest Scientific Institution Basal Research Funds(No. Y2022QC17) and the Innovation Research Program of ICT CAS (E261070).
\newpage
\section{Supplementaty Material}

\section{EM-based Hyperparameters Optimization}

The optimization process contains two parts: 1.\textbf{E-step}: Optimize $C_{ag}$; 2.\textbf{M-Step}: Optimize $w_{ms}$, $w_{rp}$, and $w_{pc}$. Here, we present a comprehensive derivation for the optimization problem proposed in the M-step.

The optimization problem in the M-step is defined by:
\begin{equation}
\begin{aligned}
\underset{w_{ms},w_{ms},w_{pc}}{\min} & C_{ag}=w_{ms}C_{ms}+w_{rp}C_{rp}+w_{pc}C_{pc}, \\
s.t. \quad
        &w_{ms}+w_{rp}+w_{pc}=1, \\
	&w_{ms}, w_{rp}, w_{pc}>\eta\\
\end{aligned}
\end{equation}
Since this optimization problem contain both equality and inequality constraints, it can be solved by utilizing the Karush–Kuhn–Tucker conditions (KKT conditions) \cite{karushMinimaFunctionsSeveral2014, kuhnNonlinearProgramming2014}. Specifically, we first reshape the problem so that it aligns with the KKT conditions:
\begin{equation}
\begin{aligned}
\underset{w_{ms},w_{ms},w_{pc}}{\min} & C_{ag}=w_{ms}C_{ms}+w_{rp}C_{rp}+w_{pc}C_{pc}, \\
s.t. \quad
        &w_{ms}+w_{rp}+w_{pc}=1, \\
	&-w_{ms}, -w_{rp}, -w_{pc}<-\eta\\
\end{aligned}
\end{equation}
Since it is not a convex optimization problem, we subsequently construct the Lagrange dual function to derive the following convex problem:
\begin{equation}
\begin{aligned}
	&L( W,\mu ,\{ \lambda _i \} ) = C_{ag}( W ) +\mu h(W) +\sum_{i=1,2,3}{\lambda _ig_i( W )}\\
	&=w_{ms}C_{ms}+w_{rp}C_{rp}+w_{pc}C_{pc}+\mu \left( w_{ms}+w_{rp}+w_{pc}-1 \right)\\
	&+\lambda _1\left( -w_{ms} + \eta \right) +\lambda _2\left( -w_{rp} + \eta \right) +\lambda _3\left( -w_{pc} + \eta \right)\\
\end{aligned}
\end{equation}
where $W=\left\{ w_{ms},w_{rp},w_{pc} \right\} $. Therefore, our objective becomes acquiring the infimum of this Lagrange dual function.
Then we decompose the above equation as follows:
\begin{equation}
\begin{aligned}
	C_{ag}\left( W \right) =&w_{ms}C_{ms}+w_{rp}C_{rp}+w_{pc}C_{pc}\\
	\mu h\left( W \right) =&\mu \left( w_{ms}+w_{rp}+w_{pc}-1 \right)\\
	\lambda _1g_1\left( W \right) =&\lambda _1\left( -w_{ms} + \eta \right)\\
	\lambda _2g_2\left( W \right) =&\lambda _2\left( -w_{rp} + \eta \right)\\
	\lambda _3g_3\left( W \right) =&\lambda _3\left( -w_{pc} + \eta \right)\\
\end{aligned}
\end{equation}
They can be solved by computing the partial derivative of each equation with respect to W and set them equal to zero:
\begin{equation}\label{equation:kkt problem}
\nabla C_{ag}\left( W^* \right) +\mu \nabla h\left( W^* \right) +\sum_{i=1,2,3}{\lambda _i\nabla g_i\left( W \right)}=0
\end{equation}
\begin{equation}\label{equation:inequ constriant}
\lambda _ig_i\left( W^* \right) = 0,i=1,2,3
\end{equation}
\begin{equation}\label{equation:equ constriant}
h\left( W^* \right) = 0
\end{equation}
\begin{equation}\label{equation:kkt constriant}
\lambda _i\ge 0,i=1,2,3
\end{equation}
\begin{equation}\label{equation:problem constriant}
g_i\left( W^* \right) \le 0,i=1,2,3
\end{equation}
In the above equations, $W^*$ is the optimal infimum of this problem. Since the problem has Slater constraint \cite{jeyakumarGeneralizationsSlaterConstraint1992} qualification, we can utilize Eq. \ref{equation:kkt problem}, Eq. \ref{equation:inequ constriant}, and Eq. \ref{equation:equ constriant} to obtain the $W^*$, and then check if the answer satisfy Eq. \ref{equation:kkt constriant} and Eq. \ref{equation:problem constriant}. The derived  $W^*$ represents the infimum of the dual problem, which is the optimal solutions for $w_{ms}$, $w_{rp}$, and $w_{pc}$.
\par

In practical experiment, since it may be challenging to obtain the analytical solution to the optimization problem in M-step, we alternatively adopt numerical optimization methods (Quasi-Newton) to obtain the optimal hyperparameters.

\section{Results on ETH3D and TnT dataset}

\begin{figure*}
    \centering
    \includegraphics[width=\linewidth]{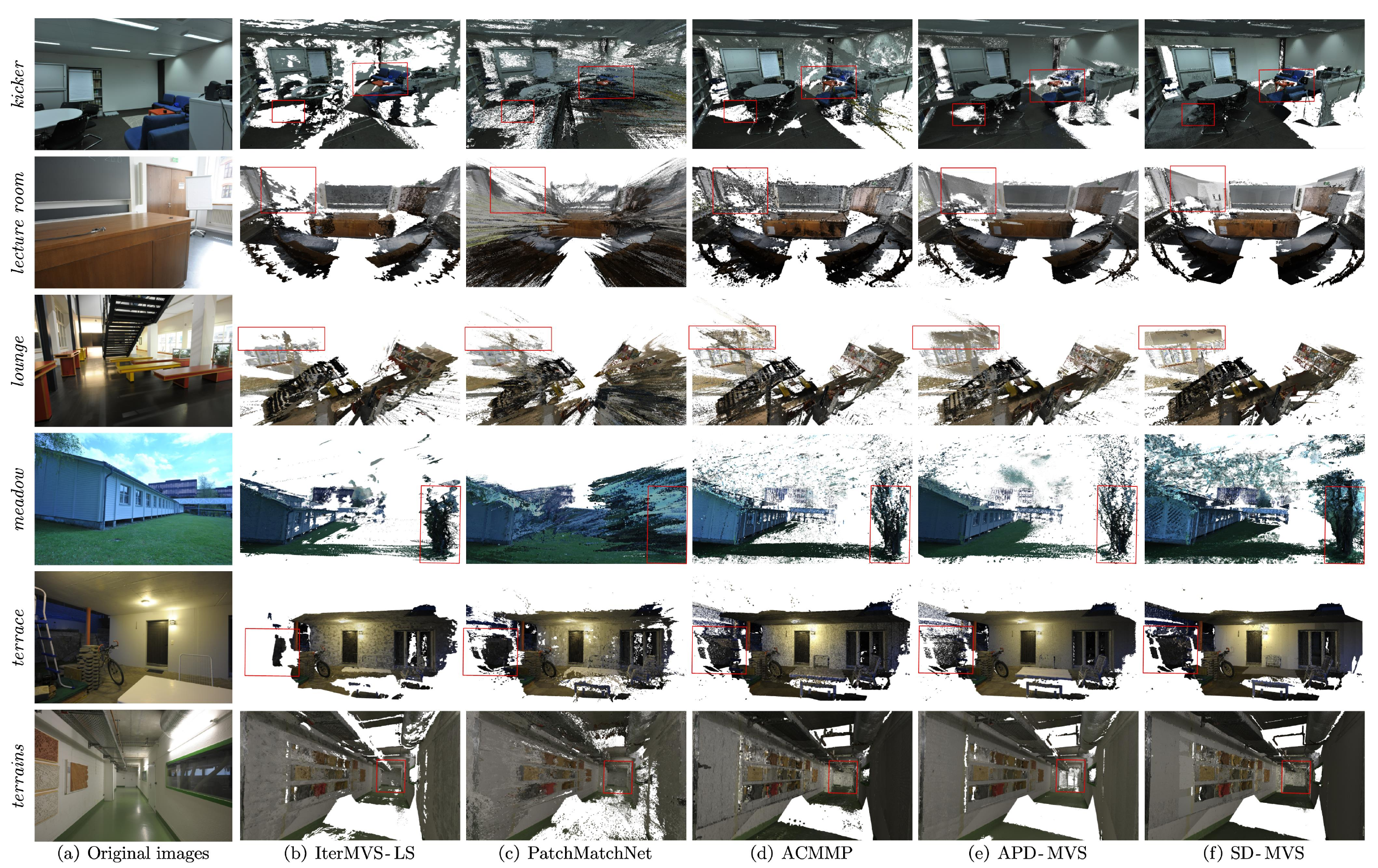}
    \caption{An illustration of the qualitative results on partial scenes of ETH3D datasets (\emph{kicker}, \emph{lecture room}, \emph{lounge}, \emph{meadow}, \emph{terrace} and \emph{terrains}). It is obvious that our methods outperform others, especially in large textureless areas.}
    \label{fig:eth3d_supp}
\end{figure*}

Fig. \ref{fig:eth3d_supp} presents some qualitative results between different methods on partial scenes of the ETH3D datasets. It is evident that our method achieves superior performance than other competing methods, especially when dealing with large textureless areas. Moreover, our method can effectively restore areas characterized by less illumination, as depicted in red boxes of \emph{meadow} and \emph{terrace}.

\section{Expanded Broader Research Context}
This section provides an expanded review of the literature to situate our work, SD-MVS, within the broader landscape of 3D computer vision, fundamental artificial intelligence methodologies, and their diverse real-world applications. While the main paper focuses on Multi-View Stereo (MVS), our research is informed by and contributes to a wider ecosystem of technological advancements, spanning from novel 3D representations and foundational models to critical applications in autonomous driving, medical imaging, and beyond.

\subsection{Advances in 3D Vision and Scene Reconstruction}

The core of our work lies in 3D reconstruction. This field has witnessed rapid progress, moving beyond traditional MVS to embrace new neural representations and robust estimation techniques.

\subsubsection{Frontiers in Multi-View Stereo (MVS)}
Our work builds upon the PatchMatch-based MVS paradigm. We have continuously explored this direction by enhancing segmentation-driven techniques with edge alignment and occlusion handling, as demonstrated in SED-MVS~\citep{yuan2025sed}, and by designing textureless-aware segmentation and refinement strategies in TSAR-MVS. Other researchers have also pushed the boundaries of MVS. For instance, Chen et al.~\citep{chen2025dual} proposed guiding MVS with dual-level precision edges for accurate planarization. The community has also investigated incorporating diverse priors, such as monocular guidance in MonoMVSNet~\citep{monomvsnet}, and adopting modern network architectures. These include leveraging Transformers for recurrent regularization in RRT-MVS~\citep{rrt-mvs} and exploring State Space Models like Mamba for improved efficiency and context modeling in MVSMamba~\citep{mvsmamba}, signaling a trend towards more powerful sequence models in 3D reconstruction.

\subsubsection{Emerging 3D Neural Representations}
Beyond MVS, neural rendering has revolutionized 3D scene representation. Neural Radiance Fields (NeRF) have set a high bar for novel view synthesis, and the NeRFBK dataset provides a holistic benchmark for evaluating such NeRF-based reconstruction methods~\citep{yan2023nerfbk}. More recently, 3D Gaussian Splatting (GS) has emerged as a highly efficient and high-quality alternative. Research in this area is flourishing, with efforts focused on creating lightweight models for dynamic 4D scenes (Light4GS)~\citep{liu2025light4gs} and deformable 2D Gaussians for real-time video representation~\citep{liu2025d2gv}. Further innovations include leveraging persistent homology to ensure topological integrity in Topology-Aware 3D Gaussian Splatting~\citep{shen2025topology}, unifying appearance codes for complex driving scenes~\citep{wang2025unifying}, and enabling controllable scene editing with 3DSceneEditor~\citep{yan20243dsceneeditor}. Directly related to our use of segmentation, GradiSeg enhances 3D boundary precision in Gaussian segmentation through gradient guidance~\citep{li2024gradiseg}. The pursuit of real-time dynamic scene rendering is also advanced by spatio-temporal decoupling techniques like STDR~\citep{li2025stdr}.

\subsubsection{Robust Depth and Stereo Estimation}
Accurate depth estimation is the cornerstone of MVS. Recent works have focused on improving robustness under challenging conditions. For example, Wang et al. have explored curriculum contrastive learning for self-supervised depth estimation in adverse weather (WeatherDepth)~\citep{Wang_2024} and have dug into using diffusion models to enhance contrastive learning for robust depth estimation~\citep{Wang_2024_2}. The power of diffusion priors is further harnessed for self-supervised depth estimation in Jasmine~\citep{wang2025jasmine}. Another line of work aims to develop a unified model that can transition from an image editor to a dense geometry estimator~\citep{wang2025editor}. In parallel, advancements in stereo matching, such as the efficient hybrid-supervised network EHSS~\citep{Zhang2023EHSSAE}, continue to improve the foundational blocks of 3D perception.

\subsection{Foundational Methods in Artificial Intelligence}

The progress in specific domains like 3D vision is heavily dependent on advancements in core AI methodologies, including foundation models, advanced segmentation, and graph learning.

\subsubsection{Vision and Language Foundation Models}
The advent of foundation models, such as the Segment Anything Model (SAM) used in our work, has transformed AI research. The field is rapidly evolving, with a focus on enhancing the capabilities of Large Language and Vision-Language Models (LLMs and VLMs). This includes promoting multi-domain reasoning through rubric-based rewards~\citep{bi2025reward}, improving model confidence on edited facts via contrastive knowledge decoding~\citep{bi2024decoding}, and aligning models for better context-faithfulness using Context-DPO~\citep{bi2024context}. Researchers are also investigating the fine-grained control of knowledge reliance by balancing parameters versus context~\citep{bi2025parameters}. On the data front, RefineX shows a path to learning how to refine pre-training data at scale~\citep{bi2025refinex}. Efficiently transitioning and scaling LLMs is addressed by methods like WISCA, which uses weight scaling~\citep{li2025wiscalightweightmodeltransition}. For VLMs, alignment is key, as explored in Re-Align, which uses retrieval-augmented DPO~\citep{xing2025re}, and DecAlign, which proposes hierarchical cross-modal alignment~\citep{qian2025decalign}. These models are being applied to complex tasks like explainable visual question answering through a diffusion chain-of-thought~\citep{lu2024explainable} and versatile advertising poster generation in AnyLayout~\citep{anonymous2025anylayout}. Their spatial intelligence is also being rigorously tested on complex reasoning benchmarks like SIRI-Bench~\citep{song2025siri}, and multimodal diffusion mamba models are unifying end-to-end generation~\citep{lu2025end}.

\subsubsection{Advanced Segmentation and Recognition}
Our "segmentation-driven" approach highlights the critical role of precise segmentation. This is a vibrant research area in its own right. For instance, in the biomedical domain, TokenUnify scales up autoregressive pretraining for neuron segmentation~\citep{chen2025tokenunify}, while multi-agent reinforcement learning is used for self-supervised neuron segmentation~\citep{chen2023self}. In general computer vision, MaskTwins introduces dual-form complementary masking for domain-adaptive segmentation~\citep{wang2025masktwins}. Novel perspectives, such as using the frequency domain, are also being explored to unlock new capabilities in medical image segmentation~\citep{han2025frequency}. For industrial applications, SSDC-Net provides an effective method for classifying steel surface defects based on salient local features~\citep{hao2024ssdc}.

\subsubsection{Graph Learning and Data Mining}
Modeling relationships and structure is crucial for many AI tasks. Graph neural networks offer a powerful framework for this. Recent research has focused on improving graph autoencoders by revisiting masking strategies from a robustness perspective~\citep{song2025equipping} and through self-purified designs like SPMGAE~\citep{song2025spmgae}. To defend against adversarial attacks, GPromptShield elevates the resilience of graph prompt tuning~\citep{song2025gpromptshield}. Furthermore, Fan et al.~\citep{fan2025multi} have demonstrated the effectiveness of multi-scale graph learning for challenging tasks like anti-sparse downscaling.

\subsubsection{Zero-Shot Learning and Prompt Engineering}
Reducing the dependency on extensive labeled data is a major goal in AI. Zero-shot learning, often powered by rich semantic information, is a promising direction. For instance, MADS leverages multi-attribute document supervision for zero-shot image classification~\citep{MADS}, while EmDepart proposes visual-semantic decomposition and partial alignment for the same task~\citep{EmDepart}. Concurrently, automating the creation of effective prompts is crucial for harnessing the power of foundation models, as explored in ProAPO for progressively automatic prompt optimization~\citep{ProAPO}.

\subsection{Applications in Diverse Domains}

The ultimate test of these technologies is their successful application in solving real-world problems. Our team and collaborators are actively engaged in deploying AI across various critical domains.

\subsubsection{Autonomous Driving and Robotics}
Autonomous driving is a primary driver for 3D vision research. Vision-Language-Action (VLA) models are becoming central, with research focusing on incentivizing reasoning and self-reflection (AutoDrive-R2)~\citep{yuan2025autodrive} and providing comprehensive surveys of pure vision-based VLA models~\citep{zhang2025pure}. End-to-end models are being simplified with distinct experts (ADDI)~\citep{zhangaddi} and made more robust via adversarial transfer (AT-Drive)~\citep{zhangdrive}. Key sub-tasks are also being addressed, such as online HD map construction with MapExpert~\citep{zhang2025mapexpert} and cross-view trajectory prediction using shared 3D queries~\citep{song2023xvtp3d}. World models with self-supervised 3D labels are being developed to enhance scene understanding~\citep{yan2025renderworld}. In robotics, physical autoregressive models show promise for manipulation without action pretraining~\citep{song2025physical}. In industry, agents like MR-IntelliAssist enable adaptive human-AI symbiosis~\citep{liu2025mr}, and cloud frameworks like A3Framework support autonomous driving path planning~\citep{10858860}.

\subsubsection{Intelligent Medical Image Analysis}
AI is revolutionizing healthcare. In medical image segmentation, a major challenge is learning from imperfect data. Significant work is being done on handling noisy labels, for instance through region uncertainty estimation~\citep{han2025region} and adaptive label correction techniques that improve robustness~\citep{qian2025adaptive}. For imbalanced data, curriculum learning frameworks like ClimD~\citep{han2025climd} and DynCIM~\citep{qian2025dyncim} are being developed to manage multimodal learning. Foundation models are also being heavily applied to pathology, with methods for fusing multi-scale heterogeneous models for whole slide image analysis~\citep{yang2025fusionmultiscaleheterogeneouspathology} and using sparse transformers for survival analysis~\citep{10226279}. In the specialized area of Cryo-Electron Tomography, self-supervised methods are used for volumetric image restoration~\citep{Yang_2021_ICCV} and denoising, guided by noise modeling and sparsity constraints~\citep{nmsg}. Approaches like Noise-Transfer2Clean~\citep{n2tc} and simulation-aware pretraining~\citep{cmb20240513} further improve denoising performance. Moreover, large models are being leveraged for generative text-guided 3D pretraining to aid segmentation, as shown in GTGM~\citep{chen2025gtgm}, and for optimizing medical prompts through evolutionary algorithms in EMPOWER~\citep{chen2025empower, 11205280}.

\subsubsection{Remote Sensing Image Interpretation}
Remote sensing provides a unique multi-view perspective of our world. Research in this area includes developing advanced multi-view graph clustering methods with dual relation optimization (MDRO)~\citep{MDRO}, structure-adaptive mechanisms (SAMVGC)~\citep{SAMVGC}, and long-short range information mining (SEC-LSRM)~\citep{SEC-LSRM}. Interactive agents like Change-Agent are being designed for comprehensive change interpretation~\citep{liu2024changeAgent}, and comprehensive surveys on spatiotemporal vision-language models for remote sensing are helping to structure the field~\citep{liu2025RSTVLM_survey}. Diffusion models are also being used for controllable remote sensing image generation (CRS-Diff)~\citep{tang2024crs} and to drive data generation for enhanced object detection in AeroGen~\citep{tang2025aerogen}.

\subsubsection{Multimodal Content Analysis and Security}
Beyond the above domains, our research extends to other areas of multimodal understanding. In composed image and video retrieval, we have explored entity mining and relation binding (ENCODER)~\citep{ENCODER}, explicit parsing of fine-grained modification semantics (FineCIR)~\citep{FineCIR}, segmentation-based focus shift revision (OFFSET)~\citep{OFFSET}, hierarchical uncertainty-aware disambiguation (HUD)~\citep{HUD}, and complementarity-guided disentanglement (PAIR)~\citep{PAIR}. In human-object interaction (HOI) detection, we are discovering syntactic interaction clues~\citep{luo2024discovering}, using context-aware instructions for multi-modal reasoning (InstructHOI)~\citep{luoinstructhoi}, and developing synergistic prompting learning frameworks~\citep{luo2025synergistic}. For video action recognition, we are exploring how to reinforce models with external tools in Video-STAR~\citep{yuan2025video}. Finally, in the domain of digital security, robust watermarking frameworks are being developed to resist extreme cropping and scaling~\citep{sunultra} and non-differentiable distortions (END2)~\citep{sun2025end}.

\bibliography{aaai24}

\end{document}